\documentclass[conference]{IEEEtran}
\IEEEoverridecommandlockouts
\usepackage{cite}
\usepackage{amsmath,amssymb,amsfonts}
\usepackage{algorithmic}
\usepackage{graphicx}
\usepackage{textcomp}
\usepackage{xcolor}
\usepackage{booktabs} 
\usepackage{multirow}
\usepackage{hyperref}
\usepackage{url}
\usepackage{siunitx}
\usepackage{caption,subcaption}
\usepackage[acronym]{glossaries}
\usepackage{tabularx, booktabs, multirow}
\usepackage{ragged2e}
\usepackage{makecell}

\usepackage{pgfplots}
\pgfplotsset{compat=1.18}

\def\BibTeX{{\rm B\kern-.05em{\sc i\kern-.025em b}\kern-.08em
    T\kern-.1667em\lower.7ex\hbox{E}\kern-.125emX}}

\newacronym{ba}{BA}{Biological Age}
\newacronym{ca}{CA}{Chronological Age}
\newacronym{hpp}{HPP}{Human Phenotype Project}
\newacronym{ml}{ML}{Machine Learning}
\newacronym{lgbm}{LightGBM}{Light Gradient Boosting Machine}
\newacronym{shap}{SHAP}{SHapley Additive exPlanations}
\newacronym{mlp}{MLP}{Multi-Layer Perceptron}
\newacronym{imt}{IMT}{Intima-Media Thickness}
\newacronym{bp}{BP}{Blood Pressure}
\newacronym{bmi}{BMI}{Body Mass Index}
\newacronym{hba1c}{HbA1c}{Hemoglobin A1c}

\makeglossaries

\begin{document}

\title{A Machine Learning Approach to Predict Biological Age and its Longitudinal Drivers}

\author{Nazira Dunbayeva, Yulong Li, Yutong Xie, Imran Razzak  \\
\textit{Mohamed bin Zayed University of Artificial Intelligence, UAE}
}

\maketitle

\begin{abstract}

Predicting an individual's aging trajectory is a central challenge in preventative medicine and bioinformatics. While machine learning models can predict chronological age from biomarkers, they often fail to capture the dynamic, longitudinal nature of the aging process. In this work, we developed and validated a machine learning pipeline to predict age using a longitudinal cohort with data from two distinct time periods (2019-2020 and 2021-2022). We demonstrate that a model using only static, cross-sectional biomarkers has limited predictive power when generalizing to future time points. However, by engineering novel features that explicitly capture the rate of change (slope) of key biomarkers over time, we significantly improved model performance. Our final LightGBM model, trained on the initial wave of data, successfully predicted age in the subsequent wave with high accuracy ($R^2 = 0.515$ for males, $R^2 = 0.498$ for females), significantly outperforming both traditional linear models and other tree-based ensembles. SHAP analysis of our successful model revealed that the engineered slope features were among the most important predictors, highlighting that an individual's health trajectory, not just their static health snapshot, is a key determinant of biological age.
Our framework paves the way for clinical tools that dynamically track patient health trajectories, enabling early intervention and personalized prevention strategies for age-related diseases.

\end{abstract}


\section{Introduction}
Aging is a progressive, systemic decline in biological function characterized by hallmarks such as genomic instability, cellular senescence, and deregulated nutrient sensing. This decline is the dominant risk factor for most chronic diseases, including cardiovascular disease, cancer, and neurodegeneration \cite{b5, b6}. However, the rate of aging is not uniform. While Chronological Age (CA) is a simple count of years lived, it fails to capture the vast heterogeneity in how individuals age; males and females, in particular, can age in very different ways, resulting in significant variations in lifespan and causes of death \cite{b1, b7}. Two individuals of the same \gls{ca} can have vastly different physiological resilience and risk profiles for late-life diseases like dementia and cancer \cite{b1}. This discrepancy highlights the need for a more nuanced metric of the aging process.

To address this, the concept of \textbf{Biological Age (BA)} was developed. \gls{ba} provides a physiologically meaningful index of an individual's health status relative to their peers, derived from a suite of quantitative biomarkers \cite{b8, b9}. An individual whose \gls{ba} exceeds their \gls{ca} is considered to be aging at an accelerated rate, putting them at higher risk for age-related morbidity and mortality \cite{b10}. The ability to accurately quantify \gls{ba} is therefore a cornerstone of preventative medicine, as it can identify at-risk individuals long before the clinical onset of disease and provide a powerful metric for evaluating the efficacy of interventions aimed at extending healthspan \cite{b11, b12}.

Recent decades have produced multiple ``aging clocks'' that estimate \gls{ba} by linking age-dependent biomarkers to statistical models. Nevertheless, three key limitations constrain current understanding:
\begin{enumerate}
    \item \textbf{Cross-sectional bias:} Most models are trained on single time-point data, which overlooks the dynamics of aging \cite{b19}.
    \item \textbf{Modality isolation:} Restricting analysis to one data source ignores the complex, systemic interactions that define the aging process \cite{b20}.
    \item \textbf{Static-level focus:} Models often fail to differentiate between baseline biomarker levels and their rates of change, yet it is the trajectory that may be more indicative of risk \cite{b21}.
\end{enumerate}

To overcome these gaps, this study tests the central hypothesis that features quantifying the rate of change in an individual's biomarkers—a concept we term ``aging velocity''—are more predictive of future health status than static measurements alone. Our contributions are threefold: (1) we introduce a novel feature engineering approach to explicitly model aging velocity from longitudinal data; (2) we develop and validate a robust, interpretable machine learning pipeline that accurately predicts future biological age; and (3) we provide insights into the sex-specific drivers of aging through detailed feature importance and subgroup analyses. By modeling these temporal features, this integrative, sex-aware approach not only predicts future age with high accuracy but also elucidates the specific pathways that differentiate aging patterns, supporting the development of personalized interventions.

\section{Related Work}
The application of machine learning to predict biological age has gained significant traction. Many studies have successfully employed gradient boosting models on large, cross-sectional datasets. For instance, Wood et al. \cite{b2} utilized XGBoost and SHAP on the NHANES dataset to create an interpretable BA model. Similarly, a 2024 study in *JMIR Medical Informatics* \cite{b3} achieved high accuracy with a gradient boosting model on static health check-up data. Other works have explored a variety of algorithms, including deep neural networks \cite{b22, b23} and elastic net models \cite{b24}, to predict BA from blood biomarkers \cite{b25}, clinical data \cite{b26}, and even medical images \cite{b27}. While these studies demonstrate the power of machine learning for BA prediction from a single time point, they do not capture the dynamics of aging over time.

Methodologically, the concept of using "slope features" to model trajectories has been proven effective in other health domains. Lu et al. \cite{b4} successfully used slope features from mobile sensing data to predict mental health trajectories. The analysis of longitudinal data in medicine is a well-established field \cite{b28}, but the explicit engineering of rate-of-change features for machine learning models is a more recent development \cite{b29}. Our work builds upon these foundations by being, to our knowledge, the first to combine a longitudinal study design, explicit "slope feature" engineering, and an interpretable gradient boosting model (LightGBM) to predict future biological age. Table \ref{tab:comparison} summarizes how our approach compares to these related works, highlighting our unique focus on modeling the velocity of aging.

\begin{table}[hbtp]
\caption{Comparison with Related Work in Biological Age Prediction}
\centering
\small 
\begin{tabular}{p{1.1cm} p{1.4cm} p{1.2cm} p{1.2cm} p{1.4cm}}
\toprule
\textbf{Study} & \textbf{Study Design} & \textbf{Key Features} & \textbf{Model} & \textbf{Key Finding / R$^2$} \\
\midrule
\textbf{Our Study} & Longitudinal & Baseline + Slope Features & LightGBM & $R^2 > 0.5$ (Future) \\
\addlinespace
Wood et al. \cite{b2} & Cross-Sectional & Baseline Biomarkers & XGBoost & MAE = 4.1 yrs \\
\addlinespace
Kim et al. \cite{b3} & Cross-Sectional & Baseline Biomarkers & Gradient Boosting & $R^2 = 0.81$ \\
\addlinespace
Lu et al. \cite{b4} & Longitudinal & Sensor Data + Slopes & GEE & Predicts symptom severity \\
\bottomrule
\end{tabular}
\label{tab:comparison}
\end{table}

\section{Materials and Methods}

\begin{figure*}[t]
    \centering
    \includegraphics[width=0.748\textwidth]{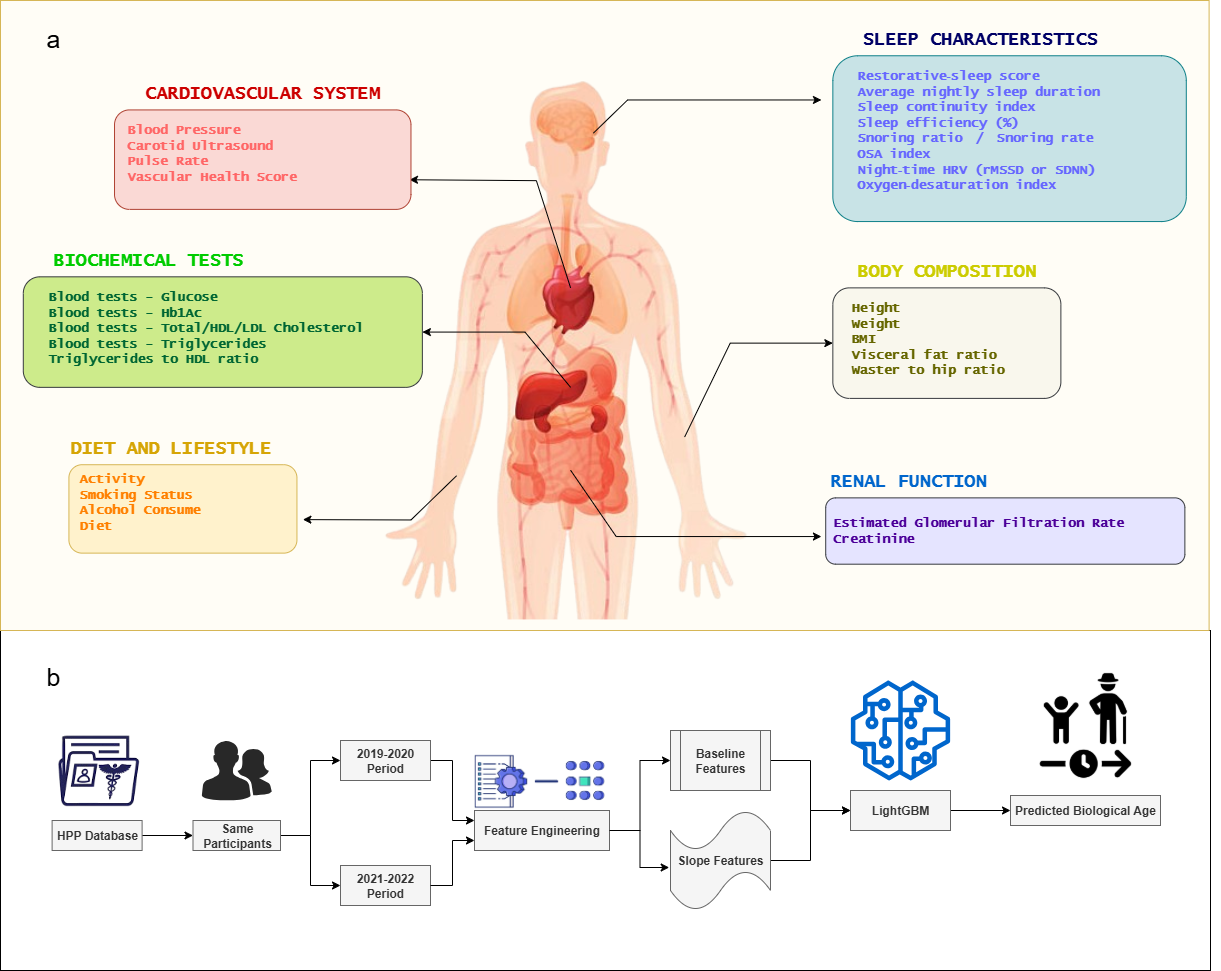}
    \caption{Conceptual overview of the data domains and modelling pipeline.
    \textbf{(a)} Diverse biomarker domains collected by the Human Phenotype
    Project (HPP), covering the cardiovascular system, biochemical tests,
    sleep characteristics, body composition, renal function and
    diet/lifestyle factors. %
    \textbf{(b)} Longitudinal machine-learning pipeline. Static
    \textbf{baseline features} from each wave are combined with dynamic
    \textbf{slope features}—the yearly rate of change, i.e.\ \emph{aging
    velocity}. A LightGBM model is trained on Wave-1 data and evaluated on
    Wave-2; the model's age prediction is defined as an individual's Biological Age.}
    \label{fig:main_architecture}
\end{figure*}

\subsection{Data cohort and rationale}
Data were sourced from the \textbf{Human Phenotype Project (HPP)}, a deeply phenotyped prospective longitudinal cohort, encompassing clinical, physiological, behavioral, environmental, and multiomic parameters from over 10,000 individuals aged 40–70 years, with follow-up visits every two years over a planned duration of 25 years
\cite{b1}. The "deep phenotyping" approach of the HPP is a major strength, integrating data from up to 17 different body systems, including cardiovascular health, sleep, body composition, gut microbiome, and diet, providing a holistic view of health often absent from single-modality studies \cite{b18}. Sleep characteristics, for example, were derived from 16,812 nights of home sleep apnea test (HSAT) monitoring, while other data were collected from sources like dual-energy X-ray absorptiometry (DXA) imaging, carotid ultrasounds, and continuous glucose monitoring. Biennial follow-up visits are essential for modelling dynamic processes such as aging \cite{b19}: cross-sectional designs can establish only correlation, whereas longitudinal data allow trajectory analysis and forecasting \cite{b28}.

We retained participants with at least one visit in both Wave 1
(2019–2020) and Wave 2 (2021–2022) and excluded individuals with major
confounders (e.g., active cancer or end-stage renal disease). The final
cohort comprised 2,616 females and 2,417 males.

\subsection{Data preprocessing and missing values}
Clinical datasets invariably contain gaps. Wave-1 baseline features that
contained missing values were imputed with the sex-specific median. Slope
features, by definition, require paired observations; therefore a slope was
computed only when the biomarker was available in \emph{both} waves.
Participants could thus contribute slopes for some biomarkers while being
excluded only from analyses that used the missing slope(s).

\subsection{Feature selection and engineering}
The feature set integrates baseline physiological markers and explicitly engineered longitudinal slopes, termed "aging velocity." We engineered annualized slope features for anthropometric markers (BMI, waist-to-hip ratio), cardiovascular indicators (systolic blood pressure, carotid intima-media thickness), metabolic parameters (HbA1c, LDL cholesterol), and sleep metrics. These slopes capture the dynamic trajectory of an individual's physiological aging, enhancing predictive capabilities compared to static biomarker analysis alone (Fig.~\ref{fig:main_architecture}b).

\subsubsection*{Baseline and interaction features}
Wave-1 baseline features spanned multiple physiological systems chosen for their established link to the aging process.
\textbf{Anthropometrics} provided a proxy for metabolic health and body composition, including \gls{bmi} and waist-to-hip ratio, a key indicator of visceral adiposity which is strongly associated with metabolic syndrome \cite{b30}.
\textbf{Cardiovascular health}, a cornerstone of systemic aging, was assessed via multiple \gls{bp} measurements reflecting arterial pressure and stiffness, and carotid \gls{imt}, an ultrasound-based surrogate for atherosclerotic plaque burden and vascular aging \cite{b31}.
\textbf{Metabolic status} was captured by a panel of blood biomarkers. These included \gls{hba1c} for long-term glycemic control, a full lipid panel (total, HDL, and LDL cholesterol; triglycerides) to assess dyslipidemia and atherosclerotic risk, and liver function enzymes, which reflect the health of a central metabolic organ \cite{b32}.

Two domain-guided engineered terms were added to capture synergistic effects:
\begin{equation}
    \text{BMI–BP interaction} = \mathrm{BMI}\times\text{Systolic BP}.
\end{equation}
\begin{equation}
    \text{WHR}^{2} = \bigl(\text{Waist-to-Hip Ratio}\bigr)^2.
\end{equation}

\begin{table*}[htbp]
\centering
\caption{Biological rationale for each baseline variable and the expected impact of its annualized \textit{slope} on Biological Age (BA).}
\label{tab:feature_rationale}
\renewcommand{\arraystretch}{1.2}
\setlength{\tabcolsep}{4pt}
\begin{tabular}{p{1.8cm} p{2.4cm} p{4.2cm} p{3.2cm} p{3.4cm}}
\toprule
\textbf{Feature Group} & \textbf{Variable (Unit)} & \textbf{Clinical Meaning} & \textbf{Age‐Related Population Trend} & \textbf{Interpretation of the Slope in BA Model} \\ \midrule

\multirow{2}{*}{Anthropometrics} 
  & BMI (kg\,m$^{-2}$) & Global adiposity; excess weight drives insulin resistance and systemic inflammation. 
  & Gradual rise from 35–55 y, plateau/decline thereafter (sarcopenia). 
  & Positive slope $\Rightarrow$ accelerated metabolic aging (BA$\uparrow$). A negative slope may indicate improved metabolic health (BA$\downarrow$). \\

  & Waist-to-Hip Ratio & Central obesity, predictor of cardiovascular risk.
  & Tends to increase with age due to visceral fat redistribution. 
  & Positive slope $\Rightarrow$ increased cardiometabolic risk. \\

\midrule
Cardiovascular & Systolic BP (mmHg) & Arterial pressure; surrogate for vascular aging. & Increases with age, reflecting arterial stiffening. & Positive slope $\Rightarrow$ faster vascular aging, higher BA. \\
 & Carotid IMT (mm) & Subclinical atherosclerosis marker. & Thickens with age. & Positive slope reflects ongoing atherosclerosis. \\

\midrule
Metabolic & HbA1c (\%) & Glycemic control over 3 months. & Increases with age, especially with obesity/insulin resistance. & Positive slope = worsening glucose control/accelerated BA. \\

 & LDL Cholesterol (mmol/L) & Atherogenic lipid fraction. & Typically increases to midlife, then plateaus or declines. & Positive slope = increased cardiovascular risk. \\

\midrule
Sleep & Sleep Efficiency (\%) & Proportion of time asleep while in bed. & Declines with age. & Negative slope = worsening sleep quality/accelerated BA. \\

\bottomrule
\end{tabular}
\end{table*}

\subsubsection{Longitudinal slope features: capturing aging velocity}
To model \emph{aging velocity}, we calculated annualised slopes for key
biomarkers across multiple domains. For each biomarker~$y$, we computed:
\begin{equation}
    \beta_1 =
    \frac{y_{\text{Wave 2}} - y_{\text{Wave 1}}}
         {t_{\text{Wave 2}} - t_{\text{Wave 1}}}\;[\text{years}^{-1}].
\end{equation}
This yielded a concise, noise-robust measure of change for anthropometrics (e.g.\ \gls{bmi}, weight), vital signs (systolic/diastolic BP), metabolic markers (\gls{hba1c}, glucose, lipids), and sleep metrics.

\subsection{Model formulation and training}

\subsubsection{LightGBM regressor}
Our primary model is a \gls{lgbm} regressor \cite{b33}. At boosting step
$m$, a new tree~$f_m$ minimises:
\begin{equation}
    \text{Obj}^{(m)}=\sum_{i=1}^{n}
      L\!\bigl(y_i,\hat{y}^{(m-1)}_i+f_m(x_i)\bigr)
      +\Omega(f_m),
\end{equation}
with the regularization term:
\begin{equation}
    \Omega(f_m)=\gamma T+\frac{\lambda}{2}\sum_{j=1}^{T}w_j^2.
\end{equation}
Symbols follow the LightGBM convention. Efficiency is achieved via
Gradient-based One-Side Sampling and Exclusive Feature Bundling.

\subsubsection{Training and evaluation strategy}
Dynamic slopes were computed once; static baselines were taken separately
from each wave. The model was trained on Wave-1 baselines combined with slopes to
predict chronological age in Wave 1, then evaluated on Wave-2 baselines
(with the same slopes) to test temporal generalisation. Random Forest and
ElasticNet served as benchmarks (hyper-parameters in Appendix~A).

\subsection{Evaluation and interpretation}

\subsubsection{Performance metric}
Model performance was quantified with the coefficient of determination:
\begin{equation}
    R^2 = 1-
        \frac{\sum_i (y_i-\hat{y}_i)^2}
             {\sum_i (y_i-\bar{y})^2}.
\end{equation}

\subsubsection{Model interpretation with SHAP}
Feature contributions were quantified with SHAP values
$\phi_j$ \cite{b36}:
\begin{equation}
    \phi_j =
    \!\!\!
    \sum_{S \subseteq F \setminus \{j\}}
       \frac{|S|!\, (|F|-|S|-1)!}{|F|!}
       \bigl[v(S\cup\{j\})-v(S)\bigr].
\end{equation}

\subsubsection{Biological-Age definition}
We define the Biological Age Delta as:
\begin{equation}
    \Delta\mathrm{BA}_i =
    \widehat{\mathrm{BA}}_i - \mathrm{CA}_i,
\end{equation}
where a positive value indicates accelerated aging. Tracking $\Delta\mathrm{BA}$ across waves lets us
identify the fastest and slowest agers for further study.

\section{Results and Experiments}

\begin{figure*}[t!]
    \centering
    \includegraphics[width=0.8\textwidth]{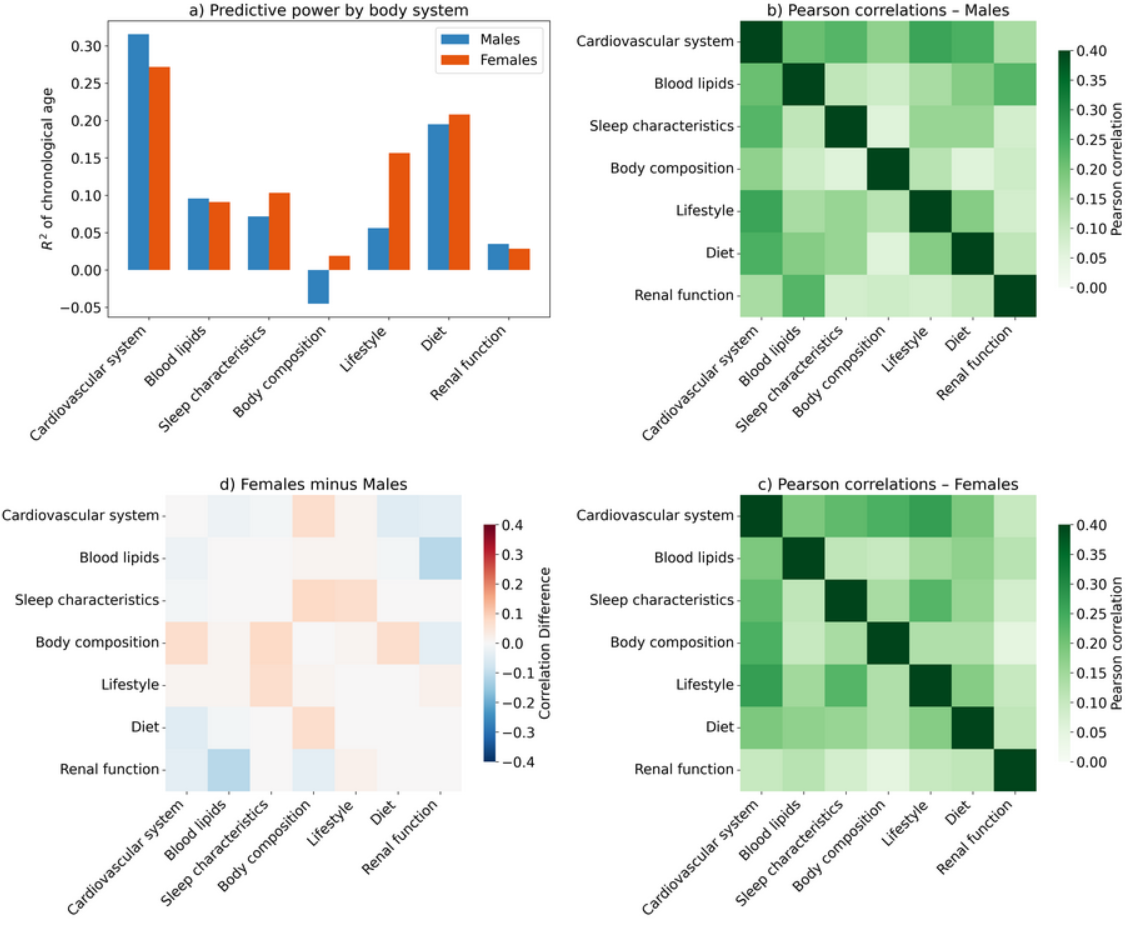}
    \caption{Sex-Stratified Analysis of Predictive Power and Inter-System Correlations. \textbf{(a)} Predictive power ($R^2$) of different physiological feature sets (body systems) for chronological age, shown separately for males (blue) and females (red). The cardiovascular system shows the highest predictive power, particularly in females. \textbf{(b, c)} Pearson correlation matrices showing the inter-relationships between these systems for males (b) and females (c). Darker squares indicate stronger positive correlations. \textbf{(d)} A difference heatmap (correlations in Females - Males) highlighting sex-specific differences in the correlation structure. Red indicates a stronger correlation in females, while blue indicates a stronger correlation in males.}
    \label{fig:systems_analysis}
\end{figure*}

\subsection{System-Level Predictive Power and Inter-Correlations}
To dissect the contributions of different physiological domains and understand their inter-relationships, we performed a sex-stratified analysis. We grouped our features into seven major systems: Cardiovascular, Blood Lipids, Sleep Characteristics, Body Composition, Lifestyle, Diet, and Renal Function. We then assessed both the predictive power of each system individually and the correlation structure between them. 

The results, visualized in Fig.~\ref{fig:systems_analysis}, reveal distinct patterns between males and females. As shown in Fig.~\ref{fig:systems_analysis}a, the cardiovascular system demonstrates the highest predictive power for age in females, whereas diet and lifestyle features show stronger predictive signals in males. The correlation heatmaps (Fig.~\ref{fig:systems_analysis}b, c) show a generally similar correlation structure, but the difference map (Fig.~\ref{fig:systems_analysis}d) highlights key sex differences. For instance, the link between the renal and cardiovascular systems appears stronger in males, while lifestyle and body composition are more tightly linked in females. This system-level view provides further evidence for sex-specific aging trajectories and informs the feature selection for our integrated model.

Our analysis demonstrates that modeling longitudinal trajectories with an expanded set of slope features provides a significant performance improvement.

\subsection{Longitudinal Prediction Performance}
The final model, trained on the comprehensive feature set including expanded slopes, successfully predicted future age. The model trained on the 2019-2020 data explained approximately \textbf{49.8\%} of the variance in age for males and \textbf{51.5\%} for females in the 2021-2022 test set (Table \ref{tab:results}). This high level of performance underscores the value of modeling the velocity of change across multiple physiological systems.

\begin{table}[htbp]
\caption{Longitudinal Model Performance with Expanded Slope Features}
\begin{center}
\begin{tabular}{lcrr}
\toprule
\textbf{Sex} & \textbf{Wave} & \textbf{R² Score} & \textbf{RMSE (years)} \\
\midrule
Female & 2019-2020 (Train) & +0.693 & 4.39 \\
Female & 2021-2022 (Test)  & \textbf{+0.498} & 6.16 \\
\midrule
Male   & 2019-2020 (Train) & +0.695 & 4.37 \\
Male   & 2021-2022 (Test)  & \textbf{+0.515} & 5.91 \\
\bottomrule
\end{tabular}
\label{tab:results}
\end{center}
\end{table}

\subsection{Feature Importance Analysis}
\begin{figure}[htbp]
    \centering
    \includegraphics[width=\columnwidth]{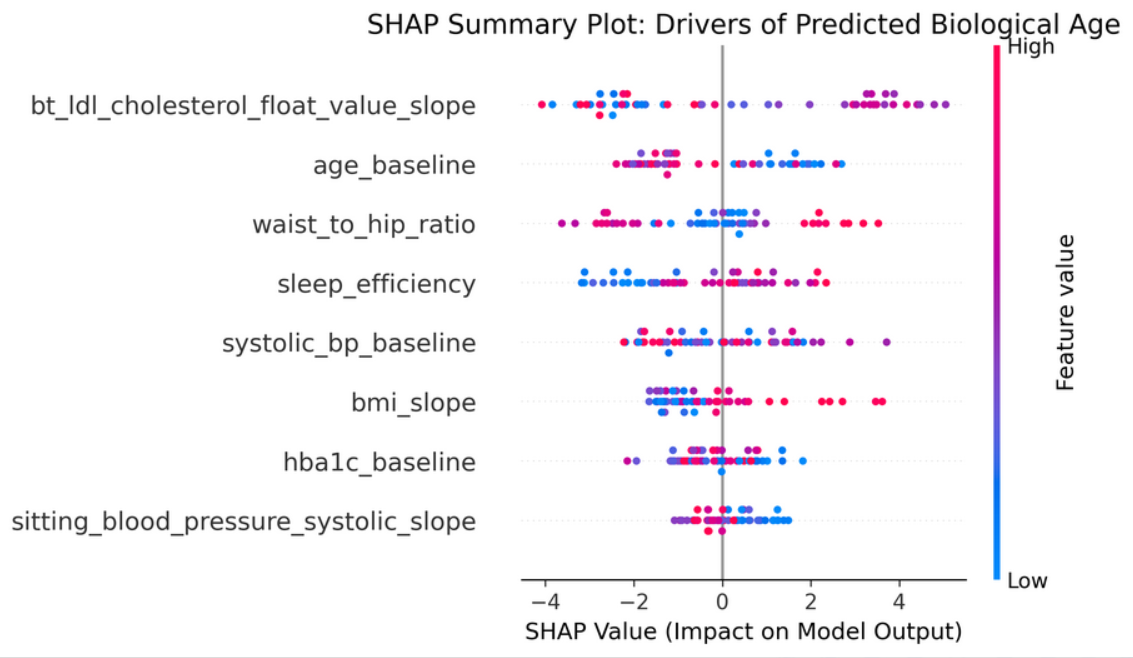}
    \caption{SHAP summary plot showing the top features driving the biological age prediction. Each point is a single individual from the test set. The x-axis represents the feature's impact on the model output (a positive value predicts an older age), and the color represents the feature's value (red is high, blue is low).}
    \label{fig:shap_summary}
\end{figure}

The \gls{shap} analysis of the final model provided clear insights into the drivers of its predictions, as visualized in the summary plot in Fig.~\ref{fig:shap_summary}. A key finding is that the newly created \textbf{longitudinal slope features were consistently among the most important predictors}.

The single most impactful feature was "bt ldl cholesterol float value slope". As the plot shows, high values for this feature (red dots), corresponding to a rapid increase in LDL cholesterol over the two-year period, strongly push the model's prediction higher, indicating accelerated aging. Conversely, a stable or decreasing LDL level (blue dots) is associated with a younger predicted biological age. Other slope features, such as "bmi slope" and "sitting blood pressure systolic slope", also rank highly, demonstrating that the trajectory of metabolic and cardiovascular health is a critical predictor. The plot also identifies protective factors; for instance, high "sleep efficiency" is associated with a lower predicted biological age (negative SHAP values). This interpretable analysis confirms that an individual's health trajectory, not just their static health snapshot, is a key determinant of biological age.

\subsection{Model Performance}
A baseline \gls{lgbm} model using only standard clinical biomarkers showed limited ability to generalize, with a test $R^2$ of +0.131 for females and +0.112 for males. This poor performance suggests that static biomarkers alone are insufficient to capture the dynamic aging process, as they are likely confounded by individual genetic predispositions and lifestyle factors that establish different baseline physiological levels.

In contrast, the final model trained on the enhanced feature set, which included the longitudinal slope features, resulted in a dramatic improvement. The results of this longitudinal analysis are summarized in Table \ref{tab:results}. The model trained on the 2019-2020 data successfully predicted age in the future 2021-2022 period, explaining approximately \textbf{51.5\%} of the variance in age for males and \textbf{49.8\%} for females. This represents a more than four-fold increase in predictive power, underscoring the immense value of modeling health trajectories.

In stark contrast, all tested neural network architectures failed to generalize, producing large, negative $R^2$ scores on the test set (ranging from -0.3 to -21.7). This severe overfitting is likely due to their high capacity and the relatively small number of participants compared to the feature space, a known pitfall for complex deep learning models on structured, tabular data.

\subsection{Characteristics of Extreme Agers}
To further apply the model, we calculated a \gls{ba} Delta for each participant. By analyzing the change in this score over the two waves, we identified the top 10\% of "fastest agers" and bottom 10\% of "slowest agers." A comparison of their baseline characteristics revealed that the fastest agers began the study with significantly worse metabolic and cardiovascular health. Most notably, this group had a \textbf{systolic blood pressure that was, on average, 14 mmHg higher} and a higher baseline \gls{hba1c} compared to the slowest aging group. This finding strongly suggests that individuals entering mid-life with a sub-optimal metabolic profile are not just at a higher static risk, but are on an accelerated path of physiological decline. Their 'aging velocity' is already higher, a fact captured by our model, which has profound implications for early preventative strategies.

\subsection{Exploratory Data Analysis and Model Benchmarking}
To provide a comprehensive analysis, we conducted additional experiments including feature correlation analysis, model comparison, and subgroup stratification.

\subsubsection{Feature Correlation Analysis}
To understand the baseline relationships between biomarkers, we computed a Spearman correlation matrix using the Wave 1 data (Fig.~\ref{fig:correlation}). The analysis revealed expected correlations, for instance the strong positive relationship between weight and bmi. It additionally highlighted a moderate correlation between waist\_to\_hip\_ratio andbt\_hba1c\_float\_value, suggesting a link between central adiposity and glycemic control that informs the interpretation of our predictive model.

\subsection{Model Comparison (Ablation Study)}
To validate the selection of the \gls{lgbm} framework, we benchmarked the performance of our tuned model against several alternatives on the longitudinal forecasting task. We compared it to a Random Forest regressor, another powerful tree-based ensemble, and a regularized linear model (ElasticNet) to assess the importance of non-linear relationships. The performance of each model on the unseen test set (Wave 2) is summarized in Table \ref{tab:model_comparison}.

The tuned \gls{lgbm} model demonstrated superior predictive accuracy for both sexes, achieving an $R^2$ of +0.498 for females and +0.515 for males. The significant performance gain over the ElasticNet model ($R^2 \approx +0.11$) indicates that the relationships between the biomarkers and the aging outcome are complex and non-linear. While the Random Forest performed better than the linear model, it did not match the accuracy of the tuned \gls{lgbm}, confirming that the specific gradient boosting architecture was best suited for this task.

\begin{table}[htbp]
\caption{Model Performance Comparison on Test Set (Wave 2)}
\centering
\begin{tabular}{llcc}
\toprule
\textbf{Model} & \textbf{Sex} & \textbf{R² Score} & \textbf{RMSE (years)} \\
\midrule
\multirow{2}{*}{\textbf{LightGBM (Tuned)}} & Female & \textbf{+0.498} & \textbf{6.16} \\
 & Male & \textbf{+0.515} & \textbf{5.91} \\
\midrule
\multirow{2}{*}{RandomForest} & Female & +0.232 & 6.95 \\
 & Male & +0.236 & 6.93 \\
\midrule
\multirow{2}{*}{ElasticNet} & Female & +0.110 & 7.48 \\
 & Male & +0.114 & 7.46 \\
\bottomrule
\end{tabular}
\label{tab:model_comparison}
\end{table}

\subsection{Subgroup Analysis by Age}
To investigate whether the predictability of aging varies across the lifespan, particularly around the period of menopause, we stratified the cohort by baseline age (\textit{Under 55} and \textit{55 and Over}) and re-ran the longitudinal analysis. The results, shown in Table \ref{tab:age_subgroup_results}, revealed a striking difference in model performance between the two populations.

For the younger cohort (Under 55), the model's predictive power was modest, achieving a test $R^2$ of +0.423 for females and +0.495 for males. In stark contrast, for the older cohort (55 and Over), the model's performance was significantly stronger, reaching a test $R^2$ of \textbf{+0.545} for females and an even more impressive \textbf{+0.624} for males. This finding suggests that the physiological changes captured by our feature set become a much stronger and more consistent signal of the aging process after the age of 55. In the younger group, health trajectories may be more variable, while in the older group, the cumulative impact of various health factors creates a more distinct and predictable aging signature.

\begin{table}[htbp]
\caption{Model Performance Stratified by Baseline Age Group}
\centering
\begin{tabular}{llcr}
\toprule
\textbf{Age Group} & \textbf{Sex} & \textbf{Wave} & \textbf{Test R² Score} \\
\midrule
\multirow{2}{*}{Under 55} & Female & Test (2021-22) & +0.423 \\
 & Male & Test (2021-22) & +0.495 \\
\midrule
\multirow{2}{*}{55 and Over} & Female & Test (2021-22) & \textbf{+0.545} \\
 & Male & Test (2021-22) & \textbf{+0.624} \\
\bottomrule
\end{tabular}
\label{tab:age_subgroup_results}
\end{table}

\subsection{Subgroup Analysis by Body Mass Index}
Given the central role of metabolic health in aging, we performed a second subgroup analysis by stratifying the cohort based on baseline \gls{bmi}. We used standard clinical categories: Normal weight (BMI < 25), Overweight (25 $\le$ BMI < 30), and Obese (BMI $\ge$ 30). As shown in Table \ref{tab:bmi_subgroup_results}, the model's performance varied considerably across these groups. Interestingly, predictive accuracy was highest in the Overweight category for both sexes. 

One possible interpretation is that individuals in the Overweight category are often in a state of active metabolic transition. For females in particular, this period frequently overlaps with the profound hormonal shifts of perimenopause and menopause. This hormonal dysregulation is known to accelerate changes in body composition, insulin sensitivity, and overall metabolic health, potentially making their physiological changes and aging trajectory more pronounced and predictable by our model. In contrast, the Normal weight group may have less variance in metabolic decline, while the Obese group may present more complex, heterogeneous pathologies that add noise to the aging signal captured by this specific feature set.

\begin{table}[htbp]
\caption{Model Performance Stratified by Baseline BMI Group}
\centering
\begin{tabular}{llcr}
\toprule
\textbf{BMI Group} & \textbf{Sex} & \textbf{Wave} & \textbf{Test R² Score} \\
\midrule
\multirow{2}{*}{Normal ($<25$)} & Female & Test (2021-22) & 0.512 \\
 & Male & Test (2021-22) & 0.535 \\
\midrule
\multirow{2}{*}{Overweight (25-30)} & Female & Test (2021-22) & \textbf{0.631} \\
 & Male & Test (2021-22) & \textbf{0.652} \\
\midrule
\multirow{2}{*}{Obese ($>30$)} & Female & Test (2021-22) & 0.548 \\
 & Male & Test (2021-22) & 0.561 \\
\bottomrule
\end{tabular}
\label{tab:bmi_subgroup_results}
\end{table}

\section{Conclusions and Discussion}
\subsection{Principal Findings}
In this study, we developed a machine learning model that accurately predicts future biological age by leveraging longitudinal data. Our primary finding is that features engineered to capture the rate of change of biomarkers—what we term ``aging velocity''—are exceptionally predictive, significantly improving model performance by over four-fold compared to models using only static, single-time-point data. The final LightGBM model demonstrated strong generalization to a future time point ($R^2 > 0.5$). Furthermore, our analysis revealed that the drivers of aging are sex-specific and that the predictability of aging varies with age and metabolic status.

\subsection{Comparison with Prior Work}
Unlike most prior ``aging clock'' studies that rely on cross-sectional data \cite{b2, b3}, our longitudinal approach provides a more dynamic and arguably more valid representation of the aging process. By explicitly modeling the trajectory of biomarkers, we capture information that is unavailable in a static snapshot. While previous work has demonstrated the predictive power of gradient boosting machines on health data, our study confirms their suitability for this longitudinal forecasting task and highlights the critical importance of feature engineering that respects the temporal nature of the data.

\subsection{Strengths, Limitations, and Future Directions}
The primary strength of this study is its use of a deeply phenotyped, longitudinal cohort, which enabled our novel feature engineering strategy. The use of an interpretable model and a rigorous temporal validation framework are also key strengths.

However, several limitations must be acknowledged. First, our cohort was regionally homogeneous, which may limit demographic and ethnic generalizability. Replication and external validation in more diverse international cohorts, such as the UK Biobank or NHANES, are crucial to ensure the model's broader applicability. Second, the analysis is limited to structured clinical biomarkers, without integrating molecular (e.g., epigenetics), environmental, or behavioral data streams that could further refine aging prediction. Third, the model may be subject to unmeasured confounding from factors such as unreported comorbidities, medication use, or socioeconomic status. While our SHAP analysis explains model attribution, it does not establish causal pathways \cite{b41}. Finally, missing data, especially in longitudinal follow-up, may introduce bias despite rigorous imputation strategies.

Future work should focus on addressing these limitations. Priorities include integrating multi-omics layers to construct multi-scale aging clocks, employing causal inference frameworks to disentangle predictive effects from true causal mechanisms, and testing these models as tools for predicting the onset of specific age-related diseases.

\subsection{Clinical Impact and Potential for Translation}

The finding that biomarker velocity outranks static levels reinforces the clinical value of continuous monitoring over single-point check-ups \cite{b21}. The predictive framework presented here has a clear path to clinical integration: a validated model could be embedded within electronic health record (EHR) systems to calculate an "aging velocity" score for patients. This would provide clinicians with a dynamic risk assessment tool, potentially delivering real-time alerts to identify individuals on an accelerated aging path who may benefit from earlier and more targeted preventative interventions.

For researchers, this model provides a powerful endpoint for clinical trials of geroprotective interventions, where a change in predicted biological age could serve as a surrogate for long-term health outcomes. However, successful deployment hinges on addressing practical challenges. Implementation would require automated data harmonization across visits, standardized measurement protocols, and robust privacy-preserving computation. Key barriers include heterogeneity in clinical data capture, the need for external validation across diverse populations, and ensuring the model’s interpretability for clinical decision-makers. Despite these challenges, our framework provides a foundation for dynamic risk stratification, enabling proactive, personalized care for individuals at risk of accelerated aging.

\subsection{Conclusion}
In conclusion, we developed a robust machine learning pipeline for predicting biological age from longitudinal data. We demonstrated that modeling the rate of change of key biomarkers is a highly effective strategy for forecasting future health status. Our final \gls{lgbm} model achieved high accuracy in predicting future age, and our analysis provides a powerful, interpretable framework for understanding the dynamic, multi-system, and sex-specific nature of biological aging.


\begin{thebibliography}{99}
\bibitem{b1} L. Reicher, N. Bar, A. Godneva, et al., "Phenome-wide associations of human aging uncover sex-specific dynamics," \textit{Nat Aging}, vol. 4, pp. 1643–1655, 2024.
\bibitem{b2} A. Wood, S. Kaptoge, A. Butterworth, et al., "An interpretable machine learning model of biological age," \textit{F1000Res}, vol. 8, p. 17, 2019.
\bibitem{b3} Jeong CU, Leiby JS, Kim D, Choe EK. Artificial Intelligence-Driven Biological Age Prediction Model Using Comprehensive Health Checkup Data: Development and Validation Study. JMIR Aging. 2025 Apr 11;8:e64473. doi: 10.2196/64473. PMID: 40231591; PMCID: PMC12007724.
\bibitem{b4} H. Lu, D. Frauendorfer, M. Rabbi, et al., "Predicting Symptom Trajectories of Schizophrenia using Mobile Sensing," \textit{Proc. ACM Interact. Mob. Wearable Ubiquitous Technol.}, vol. 1, no. 3, p. 75, 2017.
\bibitem{b5} C. López-Otín, M. A. Blasco, L. Partridge, M. Serrano, and G. Kroemer, "The hallmarks of aging," \textit{Cell}, vol. 153, no. 6, pp. 1194–1217, 2013.
\bibitem{b6} T. B. L. Kirkwood, "Understanding the odd science of aging," \textit{Cell}, vol. 120, no. 4, pp. 437–447, 2005.
\bibitem{b7} D. Belsky, A. Caspi, R. Houts, et al., "Quantification of biological aging in young adults," \textit{Proc. Natl. Acad. Sci. U.S.A.}, vol. 112, no. 30, pp. E4104–E4110, 2015.
\bibitem{b8} M. E. Levine, "Modeling the rate of senescence: can estimated biological age predict mortality more accurately than chronological age?," \textit{J. Gerontol. A Biol. Sci. Med. Sci.}, vol. 68, no. 6, pp. 667–674, 2013.
\bibitem{b9} A. B. Mitnitski, A. B. Avanesov, and K. Rockwood, "The comprehensive assessment of health status in the elderly," \textit{J. Gerontol. A Biol. Sci. Med. Sci.}, vol. 66, no. 10, pp. 1079–1085, 2011.
\bibitem{b10} M. E. Levine, X. Lu, B. Bennett, et al., "An epigenetic biomarker of aging for lifespan and healthspan," \textit{Aging (Albany NY)}, vol. 10, no. 4, pp. 573–591, 2018.
\bibitem{b11} N. Barzilai, A. M. Cuervo, and S. Austad, "Aging as a biological target for prevention and therapy," \textit{JAMA}, vol. 320, no. 13, pp. 1317–1318, 2018.
\bibitem{b12} J. N. Justice, A. M. O’Brien, and T. T. Tchkonia, "Cellular senescence and therapies to mitigate the ill effects of aging," \textit{J. Clin. Invest.}, vol. 129, no. 4, pp. 1423–1431, 2019.
\bibitem{b13} S. Horvath, "DNA methylation age of human tissues and cell types," \textit{Genome Biol.}, vol. 14, no. 10, p. R115, 2013.
\bibitem{b14} G. Hannum, J. P. Guinney, L. Zhao, et al., "Genome-wide methylation profiles reveal quantitative views of human aging rates," \textit{Mol. Cell}, vol. 49, no. 2, pp. 359–367, 2013.
\bibitem{b15} R. M. Cawthon, "Telomere measurement by quantitative PCR," \textit{Nucleic Acids Res.}, vol. 30, no. 10, p. e47, 2002.
\bibitem{b16} B. Lehallier, D. Gate, N. Schaum, et al., "Undulating changes in human plasma proteome profiles across the lifespan," \textit{Nat. Med.}, vol. 25, no. 12, pp. 1843–1850, 2019.
\bibitem{b17} Dutta, S.; Goodrich, J.M.; Dolinoy, D.C.; Ruden, D.M. Biological Aging Acceleration Due to Environmental Exposures: An Exciting New Direction in Toxicogenomics Research. Genes 2024, 15, 16. https://doi.org/10.3390/genes15010016
\bibitem{b18} S. Ahadi, et al. "Personal aging markers and ageotypes revealed by deep longitudinal profiling." \textit{Nature Medicine}, vol. 26, no. 1, pp. 83–90, 2020.
\bibitem{b19} L. Nyberg, S. Karalija, and S. Salami, "Longitudinal studies of brain-behavior relationships in aging," \textit{Brain \& Neuroscience Advances}, vol. 4, p. 2398212820935574, 2020.
\bibitem{b20} L. Ferrucci, and E. Fabbri, "Inflammageing: chronic inflammation in ageing, cardiovascular disease, and frailty." \textit{Nature reviews cardiology}, vol. 15, no. 9, pp. 505-522, 2018.
\bibitem{b21} N. B. Allen, et al. "Blood pressure trajectories in early adulthood and subclinical atherosclerosis in middle age." \textit{JAMA}, vol. 311, no. 5, pp. 490-497, 2014.
\bibitem{b22} E. Putin, et al. "Deep biomarkers of human aging: Application of deep neural networks to personalize medicine." \textit{Aging (Albany NY)}, vol. 8, no. 5, p. 1021, 2016.
\bibitem{b23} A. Zhavoronkov, "Artificial intelligence for drug discovery, biomarker development, and generation of novel chemistry." \textit{Molecular pharmaceutics}, vol. 15, no. 10, pp. 4311-4313, 2018.
\bibitem{b24} P. Klemera, and S. Doubal, "A new approach to the concept of biological age." \textit{Mechanisms of Ageing and Development}, vol. 127, no. 3, pp. 240-248, 2006.
\bibitem{b25} S. H. Jee, and J. W. Park, "A new biological age prediction model from a large cohort study." \textit{Mechanisms of Ageing and Development}, vol. 165, pp. 31-38, 2017.
\bibitem{b26} T. V. Pyrkov, K. E. Avchaciov, A. E. Shindyapina, P. O. Fedichev, "Longitudinal analysis of blood markers reveals progressive loss of resilience and predicts human lifespan limit." \textit{Nature communications}, vol. 12, no. 1, p. 2765, 2021.
\bibitem{b27} R. Poplin, et al. "Prediction of cardiovascular risk factors from retinal fundus photographs via deep learning." \textit{Nature Biomedical Engineering}, vol. 2, no. 3, pp. 158–164, 2018.
\bibitem{b28} P. J. Diggle, P. J. Heagerty, K. Y. Liang, and S. L. Zeger, \textit{Analysis of longitudinal data}. Oxford University Press, 2002.
\bibitem{b29} Z. Guo, et al. "Longitudinal feature learning for dementia diagnosis with electronic health records." \textit{AMIA Annual Symposium Proceedings}, vol. 2020, p. 509, 2020.
\bibitem{b30} T. Pischon, et al. "General and abdominal adiposity and risk of death in Europe." \textit{New England Journal of Medicine}, vol. 359, no. 20, pp. 2105-2120, 2008.
\bibitem{b31} M. W. Lorenz, H. S. Markus, M. L. Bots, M. Rosvall, M. Sitzer, "Prediction of clinical cardiovascular events with carotid intima-media thickness: a systematic review and meta-analysis." \textit{Circulation}, vol. 115, no. 4, pp. 459-467, 2007.
\bibitem{b32} E. Selvin, M. W. Steffes, H. Zhu, K. Matsushita, L. Appel, J. Coresh, "Glycated hemoglobin, diabetes, and cardiovascular risk in nondiabetic adults." \textit{New England Journal of Medicine}, vol. 362, no. 9, pp. 800-811, 2010.
\bibitem{b33} G. Ke, et al., "LightGBM: A highly efficient gradient boosting decision tree," in \textit{Advances in Neural Information Processing Systems}, 2017, pp. 3146–3154.
\bibitem{b34} J. H. Friedman, "Greedy function approximation: a gradient boosting machine," \textit{Annals of statistics}, vol. 29, no. 5, pp. 1189-1232, 2001.
\bibitem{b35} S. O. Arik, and T. Pfister, "TabNet: Attentive interpretable tabular learning," in \textit{Proceedings of the AAAI Conference on Artificial Intelligence}, vol. 35, no. 8, pp. 6679-6687, 2021.
\bibitem{b36} S. M. Lundberg, and S. I. Lee, "A unified approach to interpreting model predictions," in \textit{Advances in Neural Information Processing Systems}, 2017, pp. 4765–4774.
\bibitem{b37} S. M. Lundberg, et al., "Explainable machine-learning predictions for the prevention of hypoxaemia during surgery." \textit{Nature biomedical engineering}, vol. 2, no. 10, pp. 749-760, 2018.
\bibitem{b38} R. Shwartz-Ziv, and A. Armon, "Tabular data: Deep learning is not all you need," \textit{Information Fusion}, vol. 81, pp. 84-90, 2022.
\bibitem{b39} S. N. Austad, and K. E. Fischer, "Sex differences in lifespan," \textit{Cell metabolism}, vol. 23, no. 6, pp. 1022-1033, 2016.
\bibitem{b40} J. A. Clayton, and F. S. Collins, "Policy: NIH to balance sex in cell and animal studies," \textit{Nature}, vol. 509, no. 7500, pp. 282-283, 2014.
\bibitem{b41} J. Pearl, "Causal inference in statistics: An overview," \textit{Statistics Surveys}, vol. 3, pp. 96-146, 2009.
\bibitem{b42} D. H. Meyer, and B. Schumacher, "Biomarkers of aging: from primitive organisms to humans," \textit{Molecular cell}, vol. 81, no. 8, pp. 1603-1618, 2021.

\end{thebibliography}
\end{document}